\documentclass[letterpaper, 10 pt, conference]{ieeeconf}
\IEEEoverridecommandlockouts  %
\let\labelindent\IEEElabelindent

\usepackage[usenames,dvipsnames,svgnames,table]{xcolor}
\usepackage{amssymb,latexsym,amsfonts,amsmath}

\usepackage{bm}
\usepackage{dsfont}
\usepackage{mathtools,mathrsfs}
\usepackage{accents}
\usepackage{comment}
\usepackage{keyval}
\usepackage[normalem]{ulem}
\usepackage{graphicx}
\usepackage{epsfig}
\usepackage{tikz}
\usepackage{float} 
\usepackage{enumitem}
\usepackage{cite}
\usepackage{balance}
\usepackage[ruled, vlined]{algorithm2e}
\usepackage{booktabs}
\usepackage{multirow}
\usepackage{xcolor}

\usepackage{amsthm}
\newtheorem{thm}{Theorem}
\theoremstyle{definition}
\newtheorem{defn}{Definition}
\newtheorem{rem}{Remark}

\usepackage[colorlinks=true,linkcolor=blue,citecolor=blue,urlcolor=blue]{hyperref}

\newcommand{\R}{{\mathbb{R}}}

\newcommand{\ie}{i.e.\ }

\newcommand{\myvec}[1]{\mathbf{#1}}     %

\newcommand{\cprod}[2]{\prod_{#1}^{\raisebox{-0.5ex}{\scriptsize $#2$}}}

\newlength{\mycaption}
\newlength{\mySepBetweenFigAndCap}
\newlength{\mySepBetweenTopAndFig}
\definecolor{pink}{HTML}{ff00ff}
\definecolor{lightblue}{HTML}{3392ff}

\let\oldsum\sum
\renewcommand{\sum}{\oldsum\nolimits}
\let\oldprod\prod
\renewcommand{\prod}{\oldprod\nolimits}
\let\oldbigcap\bigcap
\renewcommand{\bigcap}{\oldbigcap\nolimits}

\title{Scalable Learning of High-Dimensional Demonstrations with\\ Composition of Linear Parameter Varying Dynamical Systems}

\author{%
    Shreenabh Agrawal$^{1,2}$,  %
    Hugo T. M. Kussaba$^{3}$, %
    Lingyun Chen$^{1}$, %
    Allen Emmanuel Binny$^{4}$, \\%
    Abdalla Swikir$^{1,5}$,
    Pushpak Jagtap$^{2}$, and
    Sami Haddadin$^{1,5}$%
\thanks{This work was funded by the German Research Foundation (DFG) as part of Germany’s Excellence Strategy, EXC 2050/1, Project ID 390696704 – Cluster of Excellence ``Centre for Tactile Internet with Human-in-the-Loop'' (CeTI) of Technische Universität Dresden. The authors also thank Martin Schonger for insightful discussions.}%
\thanks{$^{1}$Munich Institute of Robotics and Machine Intelligence (MIRMI), Technical University of Munich (TUM), Germany. %
}%
\thanks{$^{2}$Robert Bosch Center for Cyber-Physical Systems (RBCCPS), Indian Institute of Science (IISc), Bangalore, India.
}%
\thanks{$^{3}$Automation and Robotics Laboratory, Universidade de Brasília, Brazil.
}%
\thanks{$^{4}$Indian Institute of Technology (IIT), Kharagpur, India.
}%
\thanks{$^{5}$Mohamed Bin Zayed University of Artificial Intelligence (MBUZAI), Abu Dhabi, UAE.
}%
}

\begin{document}
\maketitle

\begin{abstract}
Learning from Demonstration (LfD) techniques enable robots to learn and generalize tasks from user demonstrations, eliminating the need for coding expertise among end-users. One established technique to implement LfD in robots is to encode demonstrations in a stable Dynamical System (DS).
However, finding a stable dynamical system entails solving an optimization problem with bilinear matrix inequality (BMI) constraints, a non-convex problem which, depending on the number of scalar constraints and variables, demands significant computational resources and is susceptible to numerical issues such as floating-point errors. To address these challenges, we propose a novel compositional approach that enhances the applicability and scalability of learning stable DSs with BMIs.
\end{abstract}

\section{Introduction}

Machine learning has played a key role in advancing robotics, particularly through Learning from Demonstration (LfD) techniques. These methods allow robots to learn and generalize tasks from user demonstrations, removing the need for coding expertise from end-users \cite{BillardCalinonDilmannSchall2008, argall, ravichandar2020recent, BillardMirrazaviFigueroa2022book}. A common approach to implementing LfD involves encoding demonstrations within a stable Dynamical System (DS), capturing the position and velocity of the robot's end-effector or joints during each demonstration \cite{BillardMirrazaviFigueroa2022book}. This enables robots to not only record but also generalize task learning from a few examples, empowering users to teach robots new tasks without explicit programming.

It is important to note that DSs can encode demonstrations done either in task space or joint space. Learning in task space is especially interesting when the orientation of the end-effector is not relevant for realizing the task, since the Cartesian {3D} space $\mathbb{R}^3$ is low dimensional, easing the application of learning algorithms. 
However, a DS learned in task space may produce joint-space movements that differ significantly from those in the original demonstration, as the robot can follow different joint-space paths to perform the same end-effector motion.

On the other hand, learning DSs in joint space captures better the joints interaction, which allows it to account for factors such as obstacles or joint constraints \cite{shavit2018learning}.
Additionally, learning in joint space helps avoid the need for inverse kinematics (IK) approximations, which are typically used to convert task-space motions into joint-space movements. IK-based methods often encounter issues when the robot approaches singularities, leading to erratic behavior and requiring extra engineering to maintain smooth task-space motion \cite{shavit2018learning}. %
Furthermore, as noted in \cite{shavit2018learning}, learning directly in the joint space allows reaching a task-space target in both $\mathbb{R}^3$ and $SO(3)$ without requiring explicit coupling between position and orientation \cite{ude2014orientation, kim2017gaussian, zeestraten2017approach}, which can introduce discontinuities in the generated motion. In contrast, learning in 3D task space treats the end effector as a point object and cannot capture orientation variations. This limitation is avoided when learning in the joint space (7D), as will be demonstrated in one example of Sec.~\ref{sec:simulation}.

\begin{figure}[t]
    \centering
    \includegraphics[width=\columnwidth]{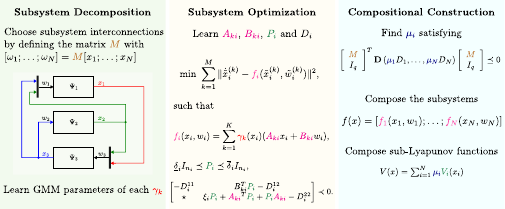}
    \caption{Our proposed approach consists of three main steps: subsystem decomposition, subsystem-level optimization and compositional construction of the DS and of the Lyapunov function certifying its stability.}
    \label{fig:visual_abstract}
\end{figure}

Nevertheless, the typical high-dimensionality of joint-space data complicates learning DSs directly from joint-space demonstrations. To the best of the author's knowledge, all DS-based LfD methods either face the challenge of solving a non-convex optimization problem or compromise accuracy to simplify the original non-convex problem into a convex optimization problem.
In particular, Linear Parameter-Varying DS (LPV-DS) based techniques such as \cite{FigueroaBillard2018PhysicalBayesianDS} and \cite{ FigueroaBillard2022Locallyactiveglobally} yield bilinear matrix inequality (BMI) constraints, making the optimization problem non-convex and generally challenging to solve from a numerical standpoint \cite{toker1995np, fukuda2001branch, roux2018validating}.
Polynomial-based techniques such as PLY-DS \cite{abyaneh2023learning} and ABC-DS \cite{SKCFSBH:24} use Sum Of Squares (SOS) based techniques which suffer from the same problem since the bilinear SOS constraints are transformed in BMI constraints.
While this problem is less evident in SEDS \cite{KhansariZadehBillard2011LearningStableNonlinear}, its representation power lags behind LPV-DS techniques \cite{FigueroaBillard2018PhysicalBayesianDS}.
Neural network-based methods \cite{kolter2019learning, rana2020euclideanizing, lawrence2020almost, abyaneh2024globally} can scale better in general due to the usage of gradient descent methods, but they are data intensive. Moreover, it is not straightforward to couple them with other beneficial restrictions to help increase the robustness of the system or decrease the overfitting (e.g., symmetrical constraints as in \cite{FigueroaBillard2022Locallyactiveglobally} or barrier certificates as in \cite{SKCFSBH:24}).

To address the challenge of scaling data-efficient techniques for learning DS from demonstrations, we propose a compositional approach that can effectively scale the LPV-DS technique proposed in \cite{FigueroaBillard2018PhysicalBayesianDS}. 
Unlike \cite{shavit2018learning}, our method does not rely on dimensionality reduction, which, while preserving essential information, can still result in the loss of critical data and potentially impact model performance.
Instead, our approach uses a divide-and-conquer strategy by (1) breaking a DS into its internal subsystems, (2) solving the DS-learning problem at the subsystem level, and (3) finding certificates that ensure the interconnection of the solutions become a valid DS.
A summary of our approach can be seen in Fig.~\ref{fig:visual_abstract}.

\section{Preliminaries} \label{prelim}

\textit{Notation}: The symbols $\mathbb{R}, \mathbb{R}_{>0}$, $\mathbb{R}_{\geq 0}$ and $\mathbb{N}$ are used to represent the sets of real numbers, positive real numbers, non-negative real numbers, and positive integers, respectively. Given $N$ vectors ${x_i \in \mathbb{R}^{n_i}}$, ${x=[x_1;...;x_N]}$ denotes the corresponding vector of dimension ${n=\sum_i n_i}$. Given a vector ${x\in\mathbb{R}^{n}}$, ${\Vert x\Vert}$ denotes the Euclidean norm of $x$. 
$I_n$ denotes the identity matrix of order $n$.
Given a symmetric matrix $A \in \mathbb{R}^{n \times n}$, the notation $A \succ 0$ denotes that $A$ is positive definite. The symbols $\prec$, $\preceq$ and $\succeq$ are defined analogously.
The $\star$ notation in a symmetric block matrix denotes the symmetric block complement.

\subsection{Learning Stable Dynamical Systems}

Let $X \subseteq \mathbb R^n$ be an open set containing the origin and consider that demonstration data in the form of $M$ tuples of position and velocity, ${({\tilde{x}}^{(k)},{{\dot{\tilde{x}}^{(k)}}})_{k=1}^{M}}$, with $\tilde{x}^{(k)} \in X$ and $\dot{\tilde{x}}^{(k)} \in \mathbb{R}^n$, is given.\footnote{%
We assume that the sequence ${(\tilde{x}^{(k)})_{k=1}^{M}}$ is made of distinct elements, and there exists a unique index $\hat{k}$ such that ${({\tilde{x}}^{(\hat{k})},{{\dot{\tilde{x}}^{(\hat{k})}}}) = (x^{*}, 0)}$.}
This demonstration data can be used to learn a stable DS 
\begin{equation}\label{eq:generic_DS}
    \dot{x}(t) = f(x(t)),
\end{equation}
with a unique equilibrium point at the origin\footnote{%
Without loss of generality, it is assumed that the desired equilibrium point ${x^*}$ is the origin, since one can use the reference data ${({\tilde{x}}^{(k)} - x^*,{{\dot{\tilde{x}}^{(k)}}})_{k=1}^{M}}$ instead of the original data, and translate the obtained DS in such way that its equilibrium point is $x^*$.%
} through an optimization problem with the function ${f\colon X \to \mathbb{R}^n}$ and a Lyapunov function ${V\colon X \to \mathbb{R}}$ as the variables of the problem~\cite{KhansariZadehBillard2011LearningStableNonlinear}. 
In particular, given parameters ${\underline \delta, \overline \delta, \xi \in \R_{> 0}}$, one can solve the following optimization problem:
\begin{subequations} \label{optimization}
\begin{equation}\label{tracking}
\min_{f} \,\, \sum_{k=1}^M \Vert {\dot{\tilde{x}}^{(k)}} - f({\tilde{x}}^{(k)}) \Vert^2,  
\end{equation}
such that
\begin{alignat}{2}
\label{fcondsys}
& f(0)=0, \\
\label{vcond1sys}
& \underline \delta \Vert x \Vert^2 \leq V(x) \leq \overline \delta\Vert x \Vert^2, && \hspace{3em} \forall x\in \mathbb{R}^{n}, \\ 
\label{vcond2sys} 
& \dot{V}(x):=\frac{\partial{V(x)}}{\partial{x}} f(x) \leq -\xi V(x), && \hspace{3em} \forall x\in \mathbb{R}^{n}.
\end{alignat}
\end{subequations}

Finding $f(x)$ and $V(x)$ for each ${x\in\mathbb{R}^{n}}$ is an infinite-dimensional optimization problem and, as such, not computationally tractable unless
a suitable parameterization of $f$ and $V$ is chosen such that
\eqref{optimization} becomes a finite-dimensional approximation.
In particular, previous approaches in the literature (cf. \cite{KhansariZadehBillard2011LearningStableNonlinear}, \cite{FigueroaBillard2018PhysicalBayesianDS}) propose a \textit{linear parameter-varying} (LPV) dynamics $f$:
\begin{equation}\label{eq:func_parameter_varying_grad_system}
    f(x) = \sum_{k=1}^K\nolimits \gamma_k (x) \, \myvec{A}_k \, x,
\end{equation}
where ${\myvec{A}_k\in\mathbb{R}^{n \times n}}$, and ${\gamma_k\colon \mathbb{R}^n\to\mathbb{R}}$ is the $K$-component Gaussian Mixture Model (GMM) defined as 
\begin{equation}\label{eq:GMM_gamma}
\gamma_k(\myvec{x}) \coloneqq \frac{\pi_k \, p(\myvec{x}|k)}{\sum_j \pi_j \, p(\myvec{x}|j)},    
\end{equation}
with ${\pi_k\ge 0}$ being the mixing weights satisfying ${\sum_{k=1}^{K} \pi_k = 1}$, and %
${p(\myvec{x}|k)}$, ${k=1,\ldots,K}$, being the probability density function of a multivariate normal distribution. These $\gamma_k$ values can be learned from demonstration data and independently of the matrices $\myvec{A}_k$  using the method prescribed in \cite{FigueroaBillard2018PhysicalBayesianDS}.
The DS corresponding to \eqref{eq:func_parameter_varying_grad_system} is given by
\begin{equation}\label{eq:LPV_DS}
    \dot{x}(t) = \sum_{k=1}^K\nolimits \gamma_k(x) \, \myvec{A}_k \, x,
\end{equation}
and the stability of the origin of \eqref{eq:LPV_DS} is certified by either (i) using the Lyapunov function ${V(x) = \| x \|^2}$, as in~\cite{KhansariZadehBillard2011LearningStableNonlinear}; or (ii) searching for a generic quadratic Lyapunov function, \ie ${V(x) = x^\top \myvec{P} x}$, where ${\myvec{P}\in\mathbb{R}^{n \times n}}$ is positive-definite, as in~\cite{FigueroaBillard2018PhysicalBayesianDS}.

\subsection{Interconnected Systems}
Instead of trying to find a stable LPV-DS by solving the large-scale optimization problem \eqref{optimization}, we will tackle multiple simpler problems by building the DS from smaller dynamical subsystems. These subsystems are defined along the dimensions of the DS, allowing us to manage complexity more effectively. The precise definitions of these subsystems, based on \cite{arcak2016networks}, are provided next.

\begin{defn}[Dynamical subsystem] \label{continuous}
	Given an index ${i\in \mathbb{N}}$, and the indexed sets ${X_i\subseteq \mathbb R^{n_i}}$, ${W_i \subseteq \mathbb R^{p_i}}$, ${Y_i \subseteq \mathbb R^{q_i}}$, a \emph{Dynamical Subsystem} (DSS) is a dynamical system  $\Psi_{i}$ characterized by the dynamics
	\begin{equation}\label{Eq_1a}
	\Psi_i\colon\left\{\hspace{-1.5mm}\begin{array}{l}\dot x_i(t) = f_i(x_i(t),w_i(t)),\\
		y_i(t)=x_i(t), \\
	\end{array}\right.
	\quad
\end{equation}
where the function ${f_i\colon X_i \times W_i \to X_i}$ satisfies the following property: there exists a unique differentiable state signal ${x_i\colon\mathbb{R}_{\geq 0} \to X_i}$ satisfying \eqref{Eq_1a} for any ${x_0 \in X_i}$ such that ${x_i(0) = x_0}$ and any \emph{internal} input signal ${w_i\colon\mathbb{R}_{\geq 0} \to W_i}$.
To simplify notation, we will also refer to a DSS $\Psi_{i}$ in \eqref{Eq_1a} by the tuple ${(X_i,W_i,f_i)}$. %
\end{defn}

These DSSs will be linked together to form an interconnected DS, which will serve as the model for the robotic task to be learned. The precise definition of the interconnection of subsystems is provided next.
\begin{defn}[Interconnection of subsystems] \label{interconnection}
	Consider $N$ dynamical sub-systems ${\Psi_i=(X_i,W_i,f_i)}$, ${i\in \{1,\dots,N\}}$,  and a matrix $M$ of an appropriate dimension defining the coupling of these subsystems.
    The \emph{interconnection} of the subsystems $\Psi_i$, denoted by ${\mathcal{I}(\Psi_1,\ldots,\Psi_N)}$, is the dynamical system $\Psi$ with dynamics given by 
    \begin{equation}\label{Eq_1a1}
        \Psi\colon \dot x(t) = f({x}(t)),
    \end{equation}
    where ${f\colon X \to \mathbb{R}^n}$, with ${X\coloneq\cprod{i=1}{N} X_i}$, is defined as 
    \begin{equation*}
        f(x) \coloneq [f_1(x_1,w_1);\dots;f_N(x_N,w_N)].
    \end{equation*} 
    where $x:=[{x_{1};\ldots\ldots;x_{N}}]$ and with the
internal variables constrained by
\begin{equation}\label{eq:M_def}
[{w_{1};\ldots\ldots;w_{N}}] = M[x_1;\ldots;x_N]. 
\end{equation}
\end{defn}

\section{Compositional Framework}\label{interconnected}

In this section, we propose a compositional framework that enables the learning of an interconnected DS ${\Psi = \mathcal{I}(\Psi_1,\ldots,\Psi_N)}$ based on its constituent DSSs ${\Psi_i, i \in \{1, \dots, N\}}$. %

To achieve this goal, we first tailor the optimization problem \eqref{optimization} for a DSS $\Psi_i$, where ${i\in\{1,\ldots,N\}}$.
Given the demonstration data ${({\tilde{x}}^{(k)},{{\dot{\tilde{x}}^{(k)}}})_{k=1}^{M}}$, we obtain new data  
${({\tilde{x}_i}^{(k)},{{\dot{\tilde{x}}_i^{(k)}}})_{k=1}^{M}}$
by projecting ${\tilde{x}}^{(k)}$ and ${\dot{\tilde{x}}^{(k)}}$ onto the set $X_i$.
The DSS-adapted version of \eqref{optimization} is the following optimization problem, where the functions ${f_i\colon X_i \times W_i \to X_i}$ and ${V_{i}\colon X_i \to \mathbb R}$ are optimization variables, %
and ${\underline \delta_i, \overline \delta_i, \xi_i \in \mathbb{R}_{> 0}}$ are optimization parameters:
\begin{subequations} \label{optimizationsubsys}
\begin{equation}\label{trackingsubsys}
\min_{f_i} \,\, \sum_{k=1}^M \Vert {{\dot{\tilde{x}}_i^{(k)}}} - f_i({\tilde{x}_i}^{(k)}, {{\tilde{w}_i}^{(k)}}) \Vert^2,  
\end{equation}
such that 
\begin{align}
    \label{fcond}
    & f_i(0, 0) = 0,  \\
    \label{vcond1}    
    & \underline \delta_i\Vert x_i \Vert^2 \leq V_i(x_i) \leq \overline \delta_i\Vert x_i \Vert^2, 
      \\
    \label{vcond2}       
    & \frac{\partial{V_i(x_i)}}{\partial{x_i}} f_i(x_i,w_i) \leq -\xi_i V_i(x_i) + \notag \\
    & \quad\quad
    \left[\begin{array}{c}w_i \\ x_i\end{array}\right]^{T} \underbrace{\left[\begin{array}{ll}D^{11}_i & D^{12}_i \\ D^{21}_i & D^{22}_i\end{array}\right]}_{D_i}\left[\begin{array}{c}w_i \\ x_i\end{array}\right],
\end{align}
\end{subequations}
where $D_i$ is a symmetric matrix of appropriate dimension with conformal block partitions $D^{m j}_i, m, j \in\{1, 2\}$. The function ${V_i\colon X_i \rightarrow \mathbb{R}}$ is called a \emph{sub-Lyapunov function} (S-LF).

\begin{rem}
During numerical optimization, it can be beneficial to add a constraint that limits the eigenvalues of the \( D_i \) matrix to ensure that they do not exceed a certain value. This helps in achieving convergence towards equilibrium.
\end{rem}

With an appropriate choice of an interconnection configuration, each optimization problem can be significantly simplified compared to the original problem \eqref{optimization}. 
However, it is important to note that the constraints \eqref{fcond}--\eqref{vcond2} alone do not guarantee the stability of each DSS.
In the following Theorem~\ref{Thm: Comp V}, we will introduce additional conditions that ensure the stability of the interconnected system 
\eqref{Eq_1a1} by combining the S-LFs from the individual DSS optimization problems into a Lyapunov function for the interconnected system.
The proof of Theorem~\ref{Thm: Comp V} is based on \cite{arcak2016networks} and is provided in Appendix~\ref{appendix_proofs}.

\begin{thm}\label{Thm: Comp V}
	Consider an interconnected system ${\Psi = \mathcal{I}(\Psi_1,\ldots,\Psi_N)}$
	arising from the composition of dynamical subsystems $\Psi_i$. Assume the following conditions hold:
	\begin{enumerate}
		\item Each subsystem $\Psi_i$ as in Definition~\ref{continuous} possesses a function $f_i$ that minimizes the objective function \eqref{trackingsubsys} and satisfies \eqref{fcond}.
		\item Each subsystem $\Psi_i$ has an S-LF $V_i$ meeting the conditions \eqref{vcond1} and \eqref{vcond2} with constants ${\vphantom{\dot{\dot{\dot{\dot{\dot{m}}}}}}\underline \delta_i, \overline \delta_i, \xi_i\in\R_{> 0}}$, and $D_i$.
		\item There exist $\mu_{i} \geq 0, i \in\{1 \ldots N\}$ such that 
		\begin{gather}\label{eq:compositional_constraint}
		{\left[\begin{array}{c}
		M \\
		I_{n}
		\end{array}\right]^{T} \mathbf{D}\left(\mu_{1} D_{1}, \ldots, \mu_{N} D_{N}\right)\left[\begin{array}{c}
		M \\
		I_{n}
		\end{array}\right] \preceq 0,} %
		\end{gather}
		where
		\begin{align}
		\notag & \mathbf{D}\left(\mu_{1} D_{1}, \ldots, \mu_{N} D_{N}\right):= \\
		& {\left[\begin{array}{llllll}
		\mu_{1} D_{1}^{11} & & & \mu_{1} D_{1}^{12} & & \\
		& \ddots & & & \ddots & \\
		& & \mu_{N} D_{N}^{11} & & & \mu_{N} D_{N}^{12} \\
		\mu_{1} D_{1}^{21} & & & \mu_{1} D_{1}^{22} & & \\
		& \ddots & & & \ddots & \\
		& & \mu_{N} D_{N}^{21} & & & \mu_{N} D_{N}^{22}
		\end{array}\right].} %
		\end{align}
	\end{enumerate}
	Then, the function ${f(x)=[f_1(x_1,w_1);\dots;f_N(x_N,w_N)]}$ minimizes the objective \eqref{tracking}, while the function $V(x)$ defined as
	\begin{equation} \label{eqV}
	    V(x) \coloneq \sum_{i=1}^N \mu_{i} V_i(x_i),
	\end{equation}
	is a Lyapunov function for the interconnected system $\Psi$ in \eqref{Eq_1a1}. 
\end{thm}

\begin{rem}
The conditions derived from Theorem~\ref{Thm: Comp V} are general and could be potentially used in the future for scaling many DS-learning algorithms that are based on \eqref{optimization}. Here, we use this Theorem~\ref{Thm: Comp V} to obtain a scalable version of LPV-DS \cite{FigueroaBillard2018PhysicalBayesianDS}.
\end{rem}
In the next theorem, we will show that in the particular case where each DSS can be written as a LPV system with internal inputs, i.e. 
\begin{equation}\label{eq:LPV-DSS}
f_i(x_i, w_i) = \sum_{k=1}^{K} \gamma_{k}(x_i)(A_{ki} x_i + B_{ki} w_i),    
\end{equation}
the constraints \eqref{vcond1} and \eqref{vcond2} of the optimization problem~\eqref{optimizationsubsys} become, respectively, a linear matrix inequality and a BMI, and it is possible to solve \eqref{optimizationsubsys} with available numerical solvers \cite{fiala2013penlabmatlabsolvernonlinear}. 
The proof of Theorem~\ref{thm:BMI_LPV_DS} is provided in the Appendix~\ref{appendix_proofs}.

\begin{thm}\label{thm:BMI_LPV_DS}
    Assume there exist constants $\underline\delta_i$, \raisebox{-0.1ex}{$\overline{\delta}_i$}, $\xi_i \in \mathbb{R}_{>0}$, and matrix variables $P_i \succ 0$, $A_{ki} \in \mathbb{R}^{n_i \times n_i}$, $B_{ki} \in \mathbb{R}^{n_i \times p_i}$, $D_i^{11} \in \mathbb{R}^{p_i \times p_i}$, $D_i^{12} \in \mathbb{R}^{p_i \times n_i}$, $D_i^{22} \in \mathbb{R}^{n_i \times n_i}$ such that the following linear and bilinear matrix inequalities hold for all $i\in\{1,\ldots,N\}$ and $k\in\{1,\ldots,K\}$:
    \begin{equation}\label{eq:LMI_eigenvalue}
        \underline\delta_i I_{n_i} \preceq P_i \preceq \overline\delta_i I_{n_i},
    \end{equation}
    \begin{equation}\label{eq:LMI_small_gain}
        \begin{bmatrix}
        -D_i^{11} & B_{ki}^T P_i - D_i^{12} \\
        \star & \xi_i P_i + A_{ki}^T P_i + P_i A_{ki} - D_i^{22}
        \end{bmatrix} \prec 0.
    \end{equation}
    Then, conditions \eqref{vcond1} and \eqref{vcond2} are satisfied.
\end{thm}

Theorems \ref{Thm: Comp V} and \ref{thm:BMI_LPV_DS} provide the foundation for Algorithm~\ref{alg:framework_optimization}, which is a scalable method for learning stable DSs from demonstrations. We first choose if we want to work in a joint space or a task space. Then, we specify a subsystem configuration to decompose the original system into an interconnected system. This choice fixes the $M$ matrix in \eqref{eq:M_def}. 
Then, for each subsystem, we fit physically-consistent GMMs \cite{FigueroaBillard2018PhysicalBayesianDS} and do optimization at a subsystem level, subject to constraints of Theorem \ref{thm:BMI_LPV_DS}, to obtain the linear DS parameters. Finally, we optimize at a global level subject to the constraints of Theorem~\ref{Thm: Comp V} to find a Lyapunov function for the interconnected DS. The GMM parameters are learned in a decoupled manner from the linear DS parameters as described in \cite{FigueroaBillard2018PhysicalBayesianDS}.
The proposed compositional framework is summarized in Fig.~\ref{fig:visual_abstract}.

\begin{algorithm}[htbp]
\caption{Proposed Framework}
\label{alg:framework_optimization}

\KwIn{Demonstrated data $\{(\tilde{x}^{(k)}, \dot{\tilde{x}}^{(k)})\}_{k=1}^M$}
\KwOut{Dynamics $f$ and Lyapunov function $V$}

\textbf{Initialize:} Shift equilibrium point to origin\;

\SetKwBlock{SubsystemDecomposition}{Subsystem Decomposition}{}
\SubsystemDecomposition{
    
    Choose subsystem configuration and specify connections defining the matrix $\textcolor{brown}{M}$:
    
    $[{w_{1};\ldots;w_{N}}]=\textcolor{brown}{M}[x_1;\ldots;x_N]$\;
    
    \For{each subsystem $\Psi_i$}{
        Obtain ${({\tilde{x}_i}^{(k)},{{\dot{\tilde{x}}_i^{(k)}}})_{k=1}^{M}}$ by projecting ${\tilde{x}}^{(k)}$ and ${\dot{\tilde{x}}^{(k)}}$ onto $X_i$.

        Learn GMM parameters of $\textcolor{Red}{\gamma_k}$ as in \cite{Mirrazavi}.\;
    }
}

\SetKwBlock{SubsystemOptimization}{Subsystem Optimization (Theorem~\ref{thm:BMI_LPV_DS})}{}
\SubsystemOptimization{
    \For{each subsystem $\Psi_i$}{

        Minimize \eqref{trackingsubsys} subject to \eqref{eq:LPV-DSS}--\eqref{eq:LMI_small_gain} 
        to find $\textcolor{magenta}{A_{ki}}$, $\textcolor{magenta}{B_{ki}}$, $\textcolor{SeaGreen}{P_i}$, $D_i^{11}$, $D_i^{12}$, $D_i^{22}$\;        
        
    }
}

\SetKwBlock{CompositionalConstruction}{Compositional Construction (Theorem~\ref{Thm: Comp V})}{}
\CompositionalConstruction{    

    Replace the numerical values of the found matrices $\textcolor{brown}{M}$ and $D_i$ in the quadratic inequality \eqref{eq:compositional_constraint} and find $\textcolor{NavyBlue}{\mu_i}$ that satisfy this constraint \;

    Let 
    
        $\textcolor{WildStrawberry}{f_i}(x_i, w_i) = \sum_{k=1}^{K} \textcolor{Red}{\gamma_{k}}(x_i)(\textcolor{magenta}{A_{ki}} x_i + \textcolor{magenta}{B_{ki}} w_i)$ \;
        
        $\textcolor{OliveGreen}{{V}_i}(x_i) = x_i^\top \textcolor{SeaGreen}{P_i} x_i$\;
    
    Define the interconnected system dynamics $f(x)$ and Lyapunov function $V(x)$ as:\;
    
    $f(x) = [\textcolor{WildStrawberry}{f_1}(x_1,w_1);\dots;\textcolor{WildStrawberry}{f_N}(x_N,w_N)]$, \;
    
    $V(x) = \sum_{i=1}^N \textcolor{NavyBlue}{\mu_{i}} \textcolor{OliveGreen}{V_i}(x_i)$
}

\end{algorithm}

\begin{rem}
While the configuration of the subsystem decomposition is not essential for the  success of the proposed method, some specific choices can yield better performance. Future investigations will focus on the development of automated tools for the optimal selection of the interconnection matrix $M$ by analyzing variance of the demonstrations within each dimension and correlations across different dimensions.
\end{rem}

\section{Simulation}\label{sec:simulation}

In this section, we present examples where stable DSs were learned from 7D joint-space data within a few minutes using our compositional approach. In all cases, the data was shifted to have the equilibrium point at the origin.
The optimization problems were implemented in MATLAB R2024a, 
utilizing the parser YALMIP \cite{lofberg2004yalmip} and the numerical solver PENLAB \cite{fiala2013penlabmatlabsolvernonlinear}.
All the simulations were done on a MacBook Pro 2022 with an Apple M2 chip, 8 GB RAM, running macOS 15.0 (24A335).

\subsection{Example 1: Individual Joint Subsystems}
\label{sub:example1_simulation}
In the first example, each joint state $q_i$ is treated as an independent subsystem, meaning $x_i = [q_i]$, where $i=1,\ldots,7$. The internal input $w_i$ for the subsystem $\Psi_i$ is the vector $[q_1;\ldots;q_{i-1};q_{i+1};\ldots;q_7]$, which consists of all joint states except $q_i$. In other words, all other joint states serve as the internal input for the subsystem $\Psi_i$, except for its own state $q_i$.

\begin{figure}[h!]
    \centering
    \includegraphics[width=0.45\textwidth]{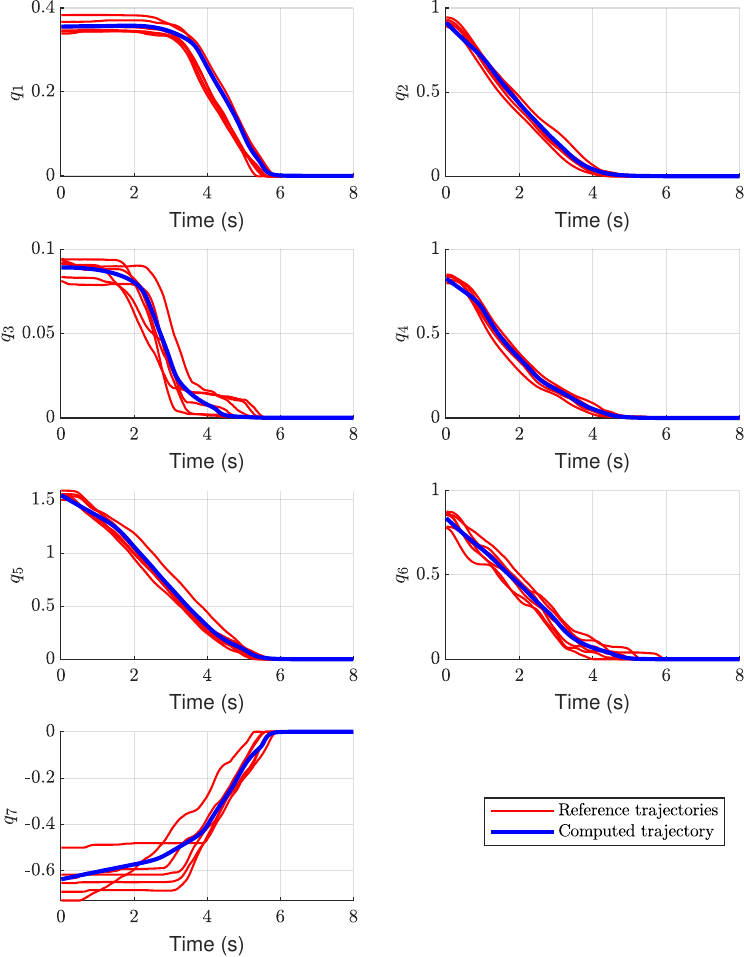}
    \caption{Comparison between demonstrated trajectories (red) in the joint-space and a generated trajectory (blue) from the learned DS using our proposed algorithm with the interconnection defined in Example 1.}
    \label{fig:joint_subsystem}
\end{figure}

As shown in Fig.~\ref{fig:joint_subsystem}, the optimizer found a solution that aligns well with the reference trajectories and satisfies all the constraints of the proposed optimization problem. The optimization process converged in a few seconds for all subsystems using the compositional approach. Specifically, the optimization took around 92 seconds, and the composition process took approximately only 9 seconds.
On the other hand, no solution was found without composition due to the solver failing to converge.
We also compare our method with SNDS, a state-of-the-art neural representation-based approach \cite{abyaneh2024globally}.\footnote{The comparison was performed using the official implementation of SNDS available at \url{https://github.com/aminabyaneh/stable-imitation-policy}.}
For the simple demonstrations of Fig.~\ref{fig:joint_subsystem},  SNDS achieved a DS with comparable mean squared error (MSE) of our compositional approach while requiring a learning time similar to our method’s running time.
A summary of the results is presented in Table~\ref{tab:comparison}.

\subsection{Example 2: Flexible Subsystem Structure}
\label{sub:example2_simulation}

\begin{figure}[h!]
    \centering
    \includegraphics[width=0.9\columnwidth]{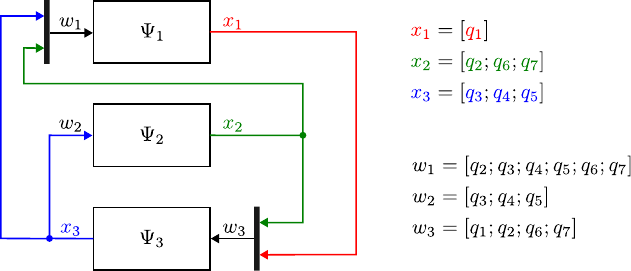}
    \caption{Interconnection between subsystems in Example 2.}
    \label{fig:3_subsys_interconnection}
\end{figure}

In the previous example, all subsystems were interconnected to each other. However, a fully connected configuration is not always necessary. %
In this second example, we demonstrate the flexibility in choosing the matrix $M$ of interconnections. Here, we use the subsystem structure specified in Fig.~\ref{fig:3_subsys_interconnection}.

\begin{table}[t!]
    \vspace*{0.55em}%
    \centering
    \caption{Comparison of mean squared error (MSE) and run time in seconds. The symbol $\times$ indicates the learning method failed to produce a stable DS due to numerical error or solver failure.}
    \begin{tabular}{l l c c}
        \toprule
        Simulation & Method & MSE & Run Time (\textit{s}) \\
        \midrule
        \multirow{3}{*}{Example 1} & \textbf{Compositional} & \textbf{2.76} & \textbf{101} \\
                                   & LPV-DS & $\times$ & $\times$ \\
                                   & SNDS (150 epochs) & 3.39 & 72.35 \\
                                   
        \midrule
        \multirow{3}{*}{Example 2} & \textbf{Compositional} & \textbf{0.37} & \textbf{146.32} \\
                                   & LPV-DS & $\times$ & $\times$ \\
                                   & SNDS (150 epochs) & $\times$ & $\times$ \\
                                   & SNDS (1000 epochs) & 7.00 & 559.66 \\
        \bottomrule
    \end{tabular}
    \label{tab:comparison}
\end{table}

\begin{figure}[h!]
    \vspace*{0.55em}%
    \centering
    \includegraphics[width=0.45\textwidth]{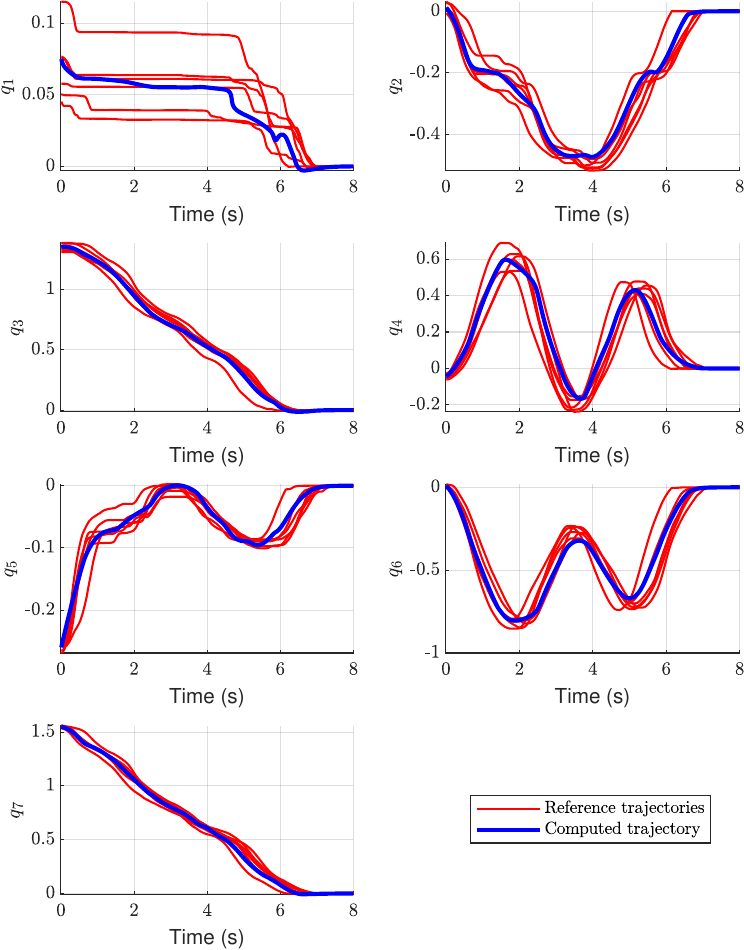}
    \caption{Comparison between demonstrated trajectories (red) in the joint-space and a generated trajectory (blue) from the learned DS using our proposed algorithm with the interconnection defined in Example 2.}
    \label{fig:flexible_subsystem}
\end{figure}

Despite the increased dimensionality of some subsystems, each optimization iteration remained fast and computationally feasible. 
The entire process of running the optimization and composition took approximately 146.36 seconds.
Solving each subsystem individually kept the number of variables and constraints manageable, whereas considering the entire problem at once in the 7D joint space led to solver failure. The solution found is plotted alongside the reference trajectories in Fig.~\ref{fig:flexible_subsystem}.
In comparison, the SNDS method \cite{abyaneh2024globally} also produced a stable solution but required a much longer training time to converge, and resulted in a higher MSE than the proposed method (see Table~\ref{tab:comparison}).

\section{Experiment}\label{sec:experiment}
In the following experiments we use a joint impedance controller to track the trajectories generated by the DSs learned in the Subsection~\ref{sec:simulation}.
The experiments are conducted with a 7-DoF Franka Emika robot~\cite{haddadin2022franka}. The robot is controlled using the Franka Control Interface (FCI) at 1 kHz using the joint velocity interface with the internal joint impedance controller.  The control loop runs on an Intel Core i5-12600K @ 4.50GHz installed with Ubuntu 22.04 and real-time kernel (6.6.44-rt39).%

\subsection{Experiment 1}

\begin{figure}[!h]
    \vspace*{0.55em}%
    \centering
    \includegraphics[width=1\columnwidth]{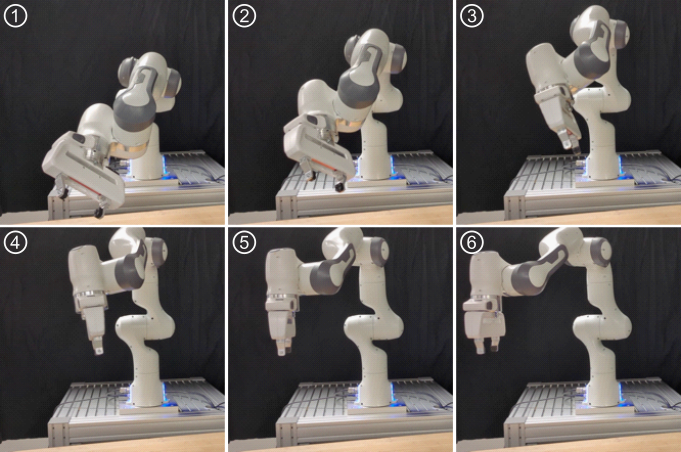}
    \caption{%
    Experimental setup corresponding to Experiment 1. The task is to move the end effector sequentially through poses \textcircled{\raisebox{-0.9pt}{1}} to \textcircled{\raisebox{-0.9pt}{6}}, matching the pose shown in the corresponding snapshots.
    }
    \label{fig:exp1_setup}
\end{figure}

\begin{figure}[!h]
    \centering
    \includegraphics[width=0.45\textwidth]{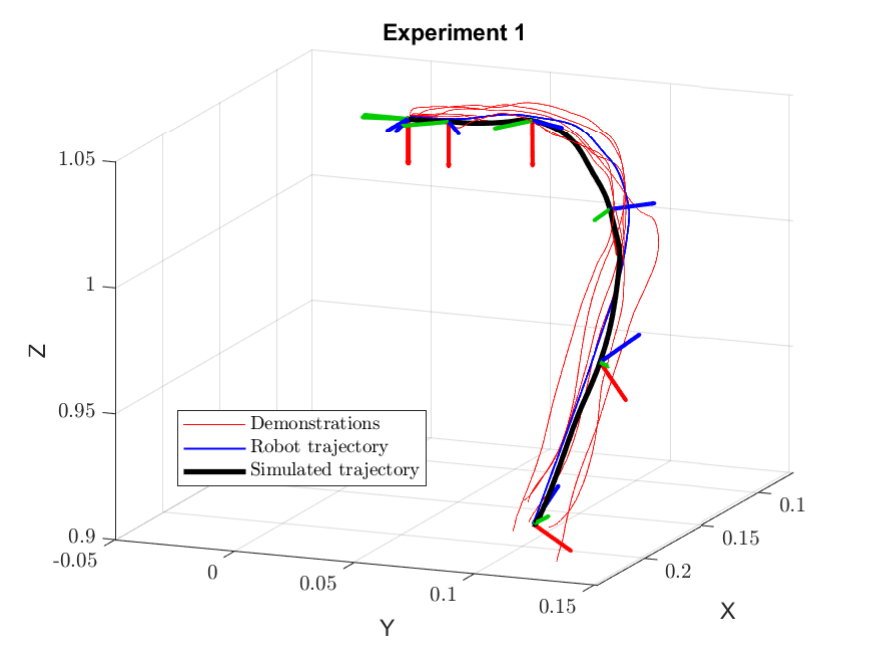}
    \caption{Reference trajectories (red), DS-generated trajectory (blue), and trajectory executed by the DS-controlled robot (black) are shown in the task space, along with orientations using RGB stick axes.}
    \label{fig:exp1_plot}
\end{figure}

The goal of this experiment is for the robot follow the motion generated by the DS learned in Subsection~\ref{sub:example1_simulation}. As shown in Fig.~\ref{fig:exp1_setup}, the desired motion involves changing the position and orientation along a non-linear path. 
In Fig.~\ref{fig:exp1_plot} we can see that the trajectory executed by the DS-controlled robot closely matches the reference demonstration trajectories, thereby validating the approach in practice.

\subsection{Experiment 2}

\begin{figure}[!h]
    \vspace*{0.55em}%
    \centering
    \includegraphics[width=1\columnwidth]{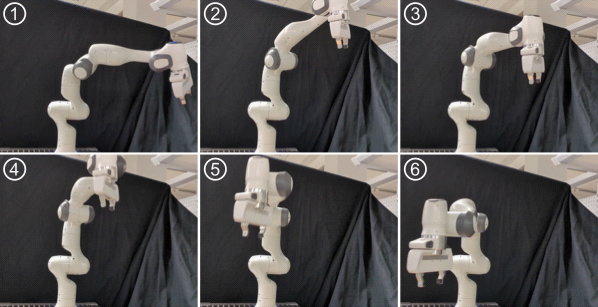}
    \caption{%
    Experimental setup corresponding to Experiment 2. The task is to move the end effector sequentially through poses \textcircled{\raisebox{-0.8pt}{1}}
 to \textcircled{\raisebox{-0.8pt}{6}}, matching the pose shown in the corresponding snapshots.
    }
    \label{fig:exp2_setup}
\end{figure}

In this experiment, we use the joint impedance controller to track the DS learned using the data set described in Subsection~\ref{sub:example2_simulation}. Although the orientation of the end-effector remains constant, the task itself is more challenging as the associated trajectories have highly-varying curvatures (see Fig.~\ref{fig:exp2_setup}). 

\begin{figure}[!h]
    \centering
    \includegraphics[width=0.45\textwidth]{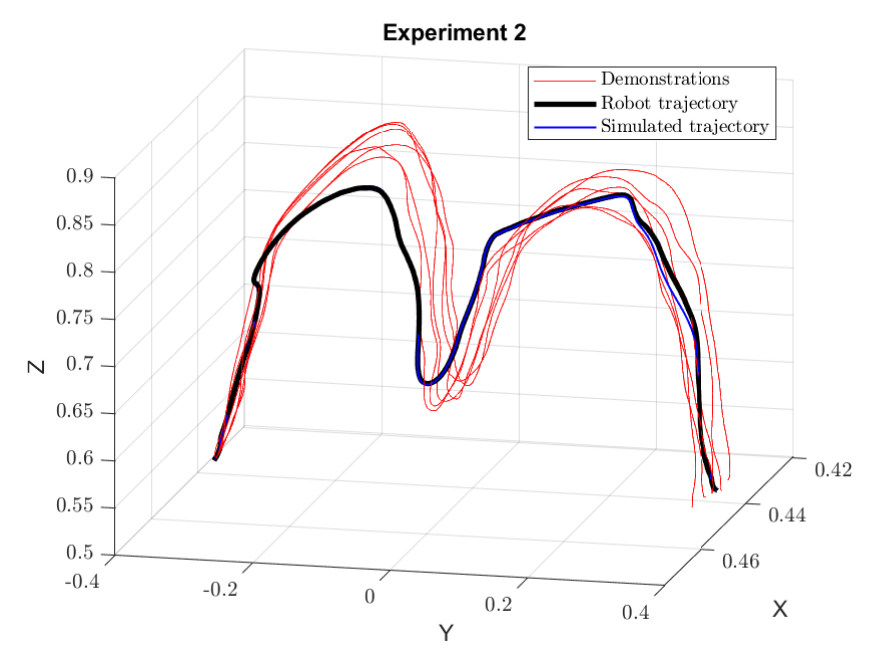}
    \caption{Reference trajectories (red), DS-generated trajectory (blue), and trajectory executed by the DS-controlled robot (black) are shown in the task space.}
    \label{fig:exp2_plot}
\end{figure}

The dataset was collected from multiple demonstrations of removing an object from inside a box without colliding with its walls \cite{SKCFSBH:24}.
As shown in Fig.~\ref{fig:exp2_plot}, the proposed compositional approach allowed learning the DS directly in the joint space, ensuring accurate replication of the demonstrations and helping to prevent collisions with the box walls.

\section{Conclusion}\label{conclusion}
This paper introduces a novel compositional method for scaling the LPV-DS technique to high-dimensional data, such as demonstrations recorded in the joint space of robot manipulators. By breaking the original optimization problem into smaller problems with fewer variables and constraints, the approach enables learning high-dimensional LPV-DSs that were previously unachievable with the vanilla technique. Moreover, despite the high dimensionality, the learning process takes only a few minutes.

\begin{appendices}

\section{}\label{appendix_proofs}

\subsection*{Proof of Theorem \ref{Thm: Comp V}}
\label{proof:Comp V}

\begin{proof}
   Using the following identity 
   \begin{align*}
    &\textstyle\sum_{k=1}^M \Vert {\dot{\tilde{x}}^{(k)}} - f({\tilde{x}}^{(k)}) \Vert^2 =\textstyle\sum_{k=1}^M \Vert {{\dot{\tilde{x}}_1^{(k)}}} - f_1(\tilde{x}_1^{(k)},\tilde{w}_1^{(k)}) \Vert^2\\
    & + \dots + \textstyle\sum_{k=1}^M \Vert {{\dot{\tilde{x}}_N^{(k)}}} - f_N(\tilde{x}_N^{(k)},\tilde{w}_N^{(k)}) \Vert^2, 
    \end{align*}
   we establish that if the functions ${f_i(x_i, w_i)}$ minimize the objective function \eqref{trackingsubsys} for all ${i \in \{1, \dots, N\}}$, then the function ${f(x) = [f_1(x_1, w_1); \dots; f_N(x_N, w_N)]}$ minimizes the objective function \eqref{tracking}.

   Moreover, it is evident that \eqref{fcond} implies  \eqref{fcondsys}. Next, we demonstrate that when \eqref{vcond1} is fulfilled, it follows that \eqref{vcond1sys} is likewise satisfied. Applying summation to both sides of the inequality \eqref{vcond1} for $i$ ranging from $1$ to $N$, one can draw the conclusion that
   \begin{align*}
   &\sum_{i=1}^N \underline \delta_i\Vert x_i \Vert^2 \leq \sum_{i=1}^N  V_i(x_i) \leq \sum_{i=1}^N \overline \delta_i\Vert x_i \Vert^2,
   \end{align*}
   which can be rewritten as
   \begin{align*}
   & \underline \delta\Vert x \Vert^2 \leq  V(x)  \leq \overline \delta\Vert x \Vert^2,
   \end{align*}
 where ${\underline \delta = N \min_{1\leq i \leq N} \underline \delta_i}$, and ${\raisebox{-0.1ex}{$\overline{\delta}$} = N \max_{1\leq i \leq N} \raisebox{-0.1ex}{$\overline{\delta}_i$}}$. Now one can show that condition \eqref{vcond2sys} holds as well, by the following chain of inequalities: 
$$
\begin{aligned}
&\dot{V}(x) = \sum_{i=1}^{N} \mu_{i} \dot{V}_{i}(x_{i}) \\
&\leq \sum_{i=1}^{N} -\mu_{i}\xi_i V_{i}(x_{i})\! +\! \mu_{i}
\left[\begin{array}{c}
w_{i} \\
x_{i}
\end{array}\right]^{T}\!\!
\left[\begin{array}{cc}
D_{i}^{11} & D_{i}^{12} \\
D_{i}^{21} & D_{i}^{22}
\end{array}\right]\!\!
\left[\begin{array}{c}
w_{i} \\
x_{i}
\end{array}\right] \\
& \leq \sum_{i=1}^{N} -\mu_{i}\xi_i V_{i}(x_{i}) + \\
&\left[\begin{array}{c}
x_{1} \\
\vdots \\
x_{N}
\end{array}\right]^{T}\!\!
\left[\begin{array}{c}
M \\
I_{n}
\end{array}\right]^{T}\!
\mathbf{D}(\mu_{1} D_{1}, \ldots, \mu_{N} D_{N})
\left[\begin{array}{c}
M \\
I_{n}
\end{array}\right]\!\!
\left[\begin{array}{c}
x_{1} \\
\vdots \\
x_{N}
\end{array}\right] \\
& \leq \sum_{i=1}^{N} -\mu_{i}\xi_i V_{i}(x_{i}).
\end{aligned}
$$
Thus, defining $\xi:=\min_{1\le i \le N}\mu_i\xi_i$, \eqref{vcond2sys} is satisfied.
\end{proof}

\subsection*{Proof of Theorem \ref{thm:BMI_LPV_DS}}
\label{proof:BMI_LPV_DS}

\begin{proof}
    Consider the quadratic Lyapunov function ${V}_i(x_i) = x_i^\top P_i x_i$. 
    Replacing $V$ in \eqref{vcond1} yields
    \[
    \underline{\delta}_i \|x_i\|^2 \leq x_i^\top P_i x_i \leq \overline{\delta}_i \|x_i\|^2, \quad \forall x_i \in X_i.
    \]
    This last inequality is true if \eqref{eq:LMI_eigenvalue} holds.
    We now prove that \eqref{eq:LMI_small_gain} implies \eqref{vcond2}. Note that \eqref{eq:LMI_small_gain} is equivalent to 
    \begin{equation}\label{eq:LMI_proof_1}
        \begin{aligned}
            &2x_{i}^{T}P_{i}A_{ki}x_{i} + 2x_{i}^{T}P_{i}B_{ki}w_{i} \le \\
            &- \xi_{i}x_{i}^{T}P_{i}x_{i} + w_{i}^{T}D_i^{11}w_{i} + 2 w_i^T D_i^{12} x_i + x_i^T D_i^{22} x_i ,
        \end{aligned}
    \end{equation}    
    for all $i=1,\ldots,N$ and $k=1,\ldots,K$. 
    Additionally, according to \eqref{eq:GMM_gamma}, we have that 
    \begin{equation}\label{eq:LMI_proof_2}
        \sum_{k=1}^K \gamma_k(x_i) = 1 \text{ and } \gamma_k(x_i) > 0     
    \end{equation}
    for each $k=1,\ldots,K$. 
    Thus, by multiplying both sides of \eqref{eq:LMI_proof_1} by $\gamma_k$ and summing both sides of the inequality for $k$ ranging from $1$ to $K$ we obtain 
    \begin{align*}
        &2x_{i}^{T}P_{i}\sum_{k=1}^{K}\gamma_{k}(x_{i})(A_{ki}x_{i}+B_{ki}w_{i}) \leq \\
        &-\xi_{i}x_{i}^{T}P_{i}x_{i} + w_{i}^{T}D_i^{11}w_{i} + 2 w_i^T D_i^{12} x_i + x_i^T D_i^{22} x_i,
    \end{align*}
    which is equivalent to \eqref{vcond2}.
\end{proof}

\end{appendices}

\balance


\end{document}